\documentclass[11pt,a4paper]{article}
\usepackage[hyperref]{emnlp2020}
\usepackage{times}
\usepackage{latexsym}
\usepackage{amsmath}
\usepackage{multirow}

\usepackage{microtype}

\usepackage{etoolbox,siunitx}	%

\robustify\bfseries
\robustify\itshape
\aclfinalcopy %
\def\aclpaperid{724} %

\usepackage{todonotes}
\usepackage{booktabs}
\usepackage{makecell}
\usepackage{listings}

\newcommand{\squishlist}{
 \begin{list}{$\bullet$}
  { \setlength{\itemsep}{0pt}
     \setlength{\parsep}{3pt}
     \setlength{\topsep}{3pt}
     \setlength{\partopsep}{0pt}
     \setlength{\leftmargin}{1.5em}
     \setlength{\labelwidth}{1em}
     \setlength{\labelsep}{0.5em} } }

\newcommand{\squishend}{
  \end{list}  }

\title{Improving QA Generalization by Concurrent Modeling of Multiple Biases} %

\author{Mingzhu Wu, Nafise Sadat Moosavi, Andreas R{\"u}ckl{\'e}, Iryna Gurevych\\ \\  Ubiquitous Knowledge Processing Lab (UKP-TUDA)\\ Department of Computer Science, Technische Universit{\"a}t Darmstadt\\ \url{https://www.ukp.tu-darmstadt.de} }

\date{}

\begin{document}
\maketitle
\begin{abstract}
Existing NLP datasets contain various biases that models can easily exploit to achieve high performances on the corresponding evaluation sets.
However, focusing on dataset-specific biases limits their ability to learn more generalizable knowledge about the task from more general data patterns.
In this paper, we investigate the impact of debiasing methods for improving generalization and propose a general framework for improving the performance on both in-domain and out-of-domain datasets by concurrent modeling of multiple biases in the training data. 
Our framework weights each example based on the biases it contains and the strength of those biases in the training data. It then uses these weights in the training objective so that the model relies less on examples with high bias weights. We extensively evaluate our framework on extractive question answering with training data from various domains with multiple biases of different strengths.  
We perform the evaluations in two different settings, in which the model is trained on a single domain or multiple domains simultaneously, and show its effectiveness in both settings compared to state-of-the-art debiasing methods.\footnote{
The code and data are available at \url{ https://github.com/UKPLab/qa-generalization-concurrent-debiasing}.}

\end{abstract}

\section{Introduction}
\label{sect:intro}

As a result of annotation artifacts, existing NLP datasets contain shallow patterns that correlate with target labels~\citep{gururangan-etal-2018-annotation,mccoy-etal-2019-right,schuster-etal-2019-towards,bras2020adversarial,Jia_2017,article}.
Models tend to exploit these shallow patterns---which we refer to as \emph{biases} 
in this paper--
instead of learning general knowledge about solving the target task.

Existing \emph{debiasing} approaches weaken the impact of such biases by disregarding or down-weighting affected training examples. %
They 
are often evaluated using adversarial or synthetic sets that contain counterexamples, in which relying on the examined bias will result in incorrect predictions \citep{belinkov-etal-2019-dont,Clark_2019,he-etal-2019-unlearn,mahabadiend}. 

Importantly, the majority of
existing debiasing approaches only deal with a single bias. %
They improve the performance scores on a targeted adversarial evaluation set, 
while typically resulting in
performance decreases on the original datasets, or on adversarial datasets that contain different types of biases~\citep{utama2020mind,nie2019analyzing,he-etal-2019-unlearn}.

In this paper, we show that modeling multiple biases is a key factor to benefit from debiasing methods for improving both in-domain performance and out-of-domain generalization, and propose a new debiasing framework for concurrent modeling of multiple biases during training.
A key challenge for developing a general framework that can handle multiple biases is to properly combine them when various biases' strength is different in each dataset. 
Previous work has found that if the ratio of biased examples is high, down-weighting, or disregarding all of them results in an insufficient training signal, which leads to performance decreases \citep{Clark_2019,utama2020mind}.
Therefore, we propose a novel multi-bias weighting function that weights each example according to multiple biases and based on each bias' strength in the training domain.
We incorporate the multi-bias weights in the training objective by adjusting the loss according to the bias weights of individual training examples so that the model relies on more general patterns of the data.

We evaluate our framework with extractive question answering (QA), for which a wide range of datasets from different domains exist---some %
contain crucial
biases \citep{weissenborn-etal-2017-making,sugawara2019assessing,Jia_2017}.

Existing approaches to improve generalization in QA %
either are only applicable when there exist
multiple %
training domains \citep{Talmor_2019,takahashi-etal-2019-cler,lee2019domainagnostic} or rely on %
models and ensembles with larger capacity
~\citep{longpre-etal-2019-exploration,su-etal-2019-generalizing,li-etal-2019-net}. %
In contrast, our novel debiasing approach can be applied to both single and multi-domain scenarios, and it
improves the model generalization 
without requiring %
larger pre-trained language models.

We compare our framework with the two state-of-the-art debiasing methods of \citet{utama2020mind} and \citet{mahabadiend}. We study its impact in two different scenarios where the model is trained on a single domain, or multiple domains simultaneously. 
Our results 
show the effectiveness of our framework compared to other debiasing methods, e.g., when the model is trained on a single domain, it improves generalization over six unseen datasets by around two points on average while the improvement is less than 0.5 points for other debiasing approaches. %

\paragraph{Our contributions:}
\begin{enumerate}
    \item We propose a new debiasing framework that handles multiple biases at once while incorporating the bias strengths in the training data. We show that the use of our framework %
    leads to
    improvements in both in-domain and out-of-domain evaluations.
        \item We are the first to investigate the impact of debiasing methods for improving generalization using multiple QA training and evaluation sets. 
\end{enumerate}

\section{Related Work}

\begin{figure*}[!htb]
	\centering
	\includegraphics[width=0.6\paperwidth]{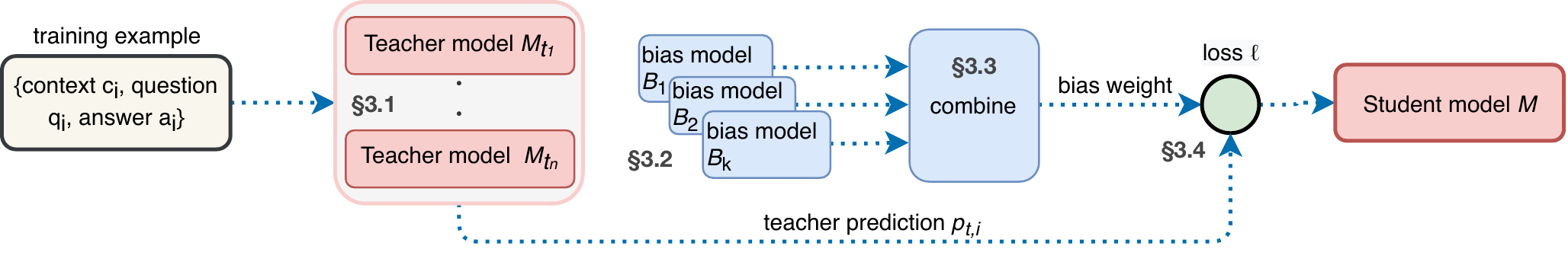}
	\caption{An illustration of our debiasing framework. The teacher and bias models are trained beforehand. During training, the corresponding teacher model for the input example outputs a prediction distribution, which will be used for distilling the knowledge to the student. Each bias model generates a bias weight for the examples. We combine all the bias weights and use them to adapt the distillation loss.}%
	\label{figure:overview}
\end{figure*}

\paragraph{Debiasing Methods}

There is a growing amount of research literature on various debiasing methods to improve the robustness of models against individual biases in the training data \citep{Clark_2019,mahabadiend,utama2020mind,he-etal-2019-unlearn,Schuster_2019}. %

The central idea of the methods proposed in previous work is to
reduce the impact of training examples that contain a bias. %
Existing work either reduces the importance of biased examples in the loss function \citep{Clark_2019,mahabadiend}, lowers the confidence on biased examples \citep{utama2020mind}, or trains an ensemble of a bias model for learning biased examples, and a base model for learning from non-biased examples \citep{Clark_2019,he-etal-2019-unlearn,mahabadiend}.

A crucial limitation of the majority of existing methods is that they only target a single bias. %
While they improve the performances on the adversarial evaluation sets crafted for this particular bias, they lead to lower performance scores on non-targeted evaluation sets including the in-domain data \citep{nie2019analyzing}, i.e., unlearning a specific bias does not indicate that the model has learned more general patterns of the data \citep{jha2020does}.
We thus need debiasing approaches that help the model to learn from less-biased patterns of the data and improve its overall performance across various datasets that are not biased or may contain different biases.

We compare our framework 
with recently proposed debiasing methods of \citet{utama2020mind} and \citet{mahabadiend}.

\citet{utama2020mind} address a single bias.
While improving the performance on the adversarial evaluation set, they also maintain the performance on the in-domain data distribution, which are exceptions to the aforementioned methods.
\citet{mahabadiend} handle multiple biases jointly and show that their debiasing methods can improve the performance across datasets if they fine-tune their debiasing methods on each target dataset to adjust the debiasing parameters.
However, the impact of their method is unclear on generalization to unseen evaluation sets.

In contrast to these state-of-the-art debiasing methods, 
we (1)~concurrently model multiple biases without requiring any information about evaluation datasets, and (2)~show that our debiasing framework achieves improvements in in-domain, as well as \emph{unseen} out-of-domain datasets. %

\paragraph{Generalization in QA} 
The ability to generalize models to unseen domains is important across a variety of QA tasks~\citep{ruckle-etal-2020-multicqa,m2020multireqa,Talmor_2019}. %
In this work, we focus on %
extractive QA.
In this context, the MRQA workshop held a shared task dedicated to evaluating the generalization capabilities of QA models %
to unseen target datasets
\cite{fisch-etal-2019-mrqa}. 
The winning team \citep{li-etal-2019-net} uses an ensemble of multiple pre-trained language models, which includes XLNet \citep{yang2019xlnet} and ERNIE \citep{sun2019ernie}. 
Other submissions outperform the baseline by using more complex models with more parameters and better pre-training. For example,
\citet{su-etal-2019-generalizing} achieve considerable improvements by simply fine-tuning XLNet instead of BERT, and \citet{longpre-etal-2019-exploration} 
achieve further improvements %
by augmenting the training data with additional unanswerable questions.

The proposed methods by \citet{takahashi-etal-2019-cler} and \citet{lee2019domainagnostic} for improving generalization leverage the fact that multiple training sets are available from different domains. For instance, \citet{takahashi-etal-2019-cler} assign an expert to each in-domain dataset, and 
\citet{lee2019domainagnostic} introduce a domain discriminator to learn domain invariant features that are shared between datasets.
Their methods are thus not applicable to a single domain scenario.

Unlike the methods mentioned above, in this paper, we propose a model-agnostic approach to handle biases of the training data for improving the generalization capability of QA models.
Our proposed approach improves 
generalization without requiring any additional training data or employing larger models or ensembles.

\section{Multi-bias Debiasing Framework}
\label{sec:methods}

Let $\mathcal{D}_T = \{D_{t_1}, \ldots D_{t_n}\}$ be the set of $n$ training datasets, and $\mathcal{D}_E = \{D_{e_{1}}, \ldots D_{e_m}\}$ 
be the set of $m$ evaluation sets that represent out-of-domain data.
Each example $x_i$ in both training and evaluation datasets contains a question $q_i$, a context $c_i$, and an answer span $a_i$ as the input. The corresponding output for $x_i$ is the start $s_i$ and end $e_i$ indices, which denote the span of the correct answer in $c_i$. %
Our goal is to train a single model on $\mathcal{D}_T$ that achieves good zero-shot transfer performances on $\mathcal{D}_E$, i.e., obtaining a generalizable model that transfers well to unseen domains.

To achieve this, we propose a novel debiasing framework that models multiple biases of the training data. 
The framework consists of four components (see Figure~\ref{figure:overview}):
(1)~multi-domain knowledge distillation (KD) to distill the knowledge from multiple teachers into a single student model (\S~\ref{subsec:kd}); (2)~a set of bias models that we use for detecting biased training examples (\S~\ref{subsec:bias_model}); (3)~a novel multi-bias weighting function that weights individual training examples based on the biases they contain (\S~\ref{subsec:mb_weighting_func}); and (4)~a bias-aware loss function, which encourages the model to focus on more general data patterns instead of heavily biases examples. We examine two different losses that either scale the teacher predictions or adjust each training example's weight during training (\S~\ref{subsec:loss_func}).

In the following, we will describe the four components in more detail.

\subsection{Multi-domain Knowledge Distillation}
\label{subsec:kd}
The idea of multi-domain knowledge distillation is to distill an ensemble of teacher models into a single student model by learning from the soft teacher labels instead of the hard one-hot labels. %
Even when only used with one training set, KD can provide a richer training signal than one-hot labels~\citep{mohebi2020,hinton2015distilling}.

We first train $n$ teacher models %
$\{M_{t_1}, \ldots, M_{t_n}\}$, one for each of the training sets. We then distill the knowledge from all the teacher models into one multi-domain student model $M$.
For every example ($x_i, y_i$) from dataset $D_j$, we obtain the probability distribution $p_{i}^{t}$ from the teacher model $M_{t_j}$ and minimize the Kullback-Leibler (KL) divergence %
between the student distribution $p_{i}^{s}$ and teacher distribution $p_{i}^{t}$. %

\subsection{Bias Models}
\label{subsec:bias_model}
In order to prevent models from learning patterns associated with biases, we first need to recognize the biased training examples.
The common method for doing so is to train models that \emph{only} leverage bias patterns
for solving the task \citep{Clark_2019,mahabadiend,utama2020mind,he-etal-2019-unlearn}. We call these models \emph{bias models} $B_1, \ldots, B_k$.
For instance, some answers can be identified %
by only considering the interrogative adverbs that indicate the question types, e.g., \textit{when}, \textit{where}, etc. \citep{weissenborn-etal-2017-making}. 
Therefore, the corresponding bias model will only uses those adverbs in questions to %
identify answers. 

We use such bias models to compute weights that determine how %
well the training examples can be solved by relying on the biases.

Since QA models should predict the indices of the start and end tokens of an answer span, we define two bias weights $\beta_{j,s}$ and $\beta_{j,e}$ for each example $x_i$. 
Assuming $B_j(x_i) = \{b_1, \ldots, b_{|c_i|}\}$ is $B_j$'s predicted output distribution of the start index for $x_i$ and $g$ is the gold start index, we define $\beta_{j,s}$ as follows:
\begin{equation}
  \beta_{j,s}(x_i) =
    \begin{cases}
      b_{g} & \text{if the prediction is correct}\\
      0 & \text{otherwise}
    \end{cases}   \nonumber   
\end{equation}
where the start index prediction of $B_j$ on $x_i$ is correct if $\mathit{argmax}(B_j(x_i)) = g$.
By setting $\beta_{j,s}$ to zero, we treat the example as unbiased if it cannot be answered by the bias model. 

We determine $\beta_{j,e}$ accordingly for the end index. To simplify our notation, in the remainder of this work, we denote $\beta(x_i)$ as the bias weight of one example and do not differentiate between start and end indices. %

\subsection{Multi-Bias Weighting Function}
\label{subsec:mb_weighting_func}

As we show in \S~\ref{sec:bias_strength}, each dataset contains various biases with different strengths. 
If we directly use the output of the bias models to down-weight or filter all biased examples, as it is the case in existing debiasing methods, we will lose the training signal from a considerable portion of the training data. This will in turn decrease the overall performance~\cite{utama2020mind}.
To apply our framework to training sets that may contain multiple biases of different strengths, %
we automatically weight the output of the bias models according to the strength of each bias in each training dataset.

Therefore, we propose a scaling factor $F_S(B_k,D_{t_j})$ to automatically control the impact of bias $B_k$ 
in dataset $D_{t_j}$ in our debiasing framework, i.e., to reduce the impact of bias on the loss function when the bias is commonly observed in the dataset. 

The scaling factor is defined as:
\begin{equation}
    F_S(B_k,D_{t_j})= 1 - \frac{\text{EM}(B_k,D_{t_j})}{\text{EM}(M_{t_j},D_{t_j})} \label{eq:scaling-factor}
\end{equation}
where EM measures the performance of the examined model on the given dataset based on the exact match score, and $M_{t_j}$ is the teacher model that is trained on $D_{t_j}$. 
This lowers the impact of strong biases whose corresponding bias models perform well, e.g., when their performance is close to the performance of the teacher model.
If $F_S = 0$, the performance of $B_k$ equals to $M_{t_j}$, indicating that this bias type exists in all the training examples. Thus, we do not use it for debiasing.

We then combine multiple biases %
for a single training example $x_i \in D_{t_j}$ as follows:
\begin{equation}
    F_B (x_i) =  \min_k(F_S(B_k, D_{t_j}) \times \beta_{k}(x_i)) 
    \label{eq:combine}
\end{equation}
The scaling factor $F_S(B_k,D_{t_j})$ computes a \emph{dataset-level} weight for bias $B_k$ while $\beta_{k}(x_i)$ computes an \emph{example-level} weight for $x_i$ based on $B_k$. 
In summary, an example $x_i$ receives a high weight based on $B_k$ if (1) $x_i$ can be correctly answered using the bias model $B_k$, and (2) $B_k$ is not prevalent in the training examples of $D_{t_j}$.
The final bias weight $F_B (x_i)$ of a bias $B_k$ on example $x_i$ is the product of the example-level and dataset-level weights.

The purpose of using the minimum in Equation~\ref{eq:combine} is to retain as much training signal as possible from the original data by only down-weighting examples that are affected by all biases.

\subsection{Bias-Aware Loss Function}
\label{subsec:loss_func}
The final step is to incorporate $F_B$ within
the distillation process to adapt the loss of each example based on its corresponding bias weight.

Assume $p_{i}^{t}$ and $p_{i}^{s}$ are the probability predictions of a teacher model $M_{t_j}$ and a student model $M$ on example $x_i \in D_{t_j}$, respectively. 
We incorporate $F_B$ in the loss function in two different ways: (1) multi-bias confidence regularization (\emph{Mb-CR}), and (2) multi-bias weighted loss (\emph{Mb-WL}).
While bias weights are used to scale the teacher probabilities in \emph{Mb-CR}, they are %
directly applied to weight the training loss in \emph{Mb-WL}.
The main difference between these two training losses is that the bias weights have a more direct and therefore a stronger impact on the loss function in \emph{Mb-WL}.

\paragraph{Multi-bias confidence regularization (\emph{Mb-CR}).}
We adapt the confidence regularization method of \citet{utama2020mind} to our setup to concurrently debias multiple biases. %
We use $F_B$ to scale the teacher predictions to make the teacher
less confident on biased examples. We define the scaled %
probability of the teacher model on token $j$ of $x_i$ as follows: %
\begin{equation}
    S(p_{i}^{t}, F_B(x_i))_j = \frac {p_{i,j}^{(1-F_B(x_{i}))}} {\sum_{k} p_{i,k}^{(1-F_B(x_{i}))}}   
\end{equation}
We then train the student model $M$ by minimizing the Kullback-Leibler divergence 
 between $p_{i}^{s}$ and $S(p_{i}^{t}, F_B(x_i))$:\footnote{The final loss is the average of the start and end losses, which are both computed using the same loss function $\mathcal{L}$.}
\begin{equation} 
\begin{aligned}
\mathcal{L}(x_i, S(p_{i}^{t}, F_B(x_i))) =  \text{KL}(\log p_{i}^{s}, S(p_{i}^{t}, F_B(x_i))) \nonumber
\end{aligned}
\end{equation}

\paragraph{Multi-bias weighted loss (\emph{Mb-WL}).}
In this approach, we use the bias weights to directly weight the corresponding loss of each training example.
In this case, the training objective is to minimize the weighted Kullback-Leibler divergence $\mathcal{L}$ between $p_{i}^{s}$ and $p_{i}^{t}$ as follows: 
\begin{equation} 
\label{alternative_score} 
\begin{aligned}
\mathcal{L}(x_i) = & (1- F_B(x_i)) \times \text{KL}(\log p_{i}^{s}, p_{i}^{t}) \nonumber
\end{aligned}
\end{equation}

\section{Experimental Setup}

\subsection{Base Model}
We perform all experiments with BERT base uncased~\cite{devlin-etal-2019-bert} in the AllenNLP framework~\cite{Gardner2017AllenNLP}.
We use the MRQA multi-task  implementation~\cite{fisch2019mrqa} of BERT for QA model as the baseline. %

\subsection{Examined Biases and Bias Models}
\label{sect:bias_models}
We incorporate four biases in our experiments. 

\squishlist
\item \emph{Wh-word} \citep{weissenborn-etal-2017-making}: %
the corresponding model for detecting this bias 
only uses the interrogative adverbs from the question.
\item \emph{Lexical overlap} \citep{Jia_2017}: 
in many QA examples, the answer is in the sentence of the context that has a high similarity to the question. To recognize this bias, we train the bias model using only the sentence of the context that has the highest similarity to the question, if the answer lies in this sentence.\footnote{We use Sentence-BERT \citep{reimers-gurevych-2019-sentence} to determine the sentence similarity.}  Otherwise, we exclude the example during training. 

\item \emph{Empty question} \citep{sugawara2019assessing}: 
the answer can be found without the presence of a question, e.g., by selecting the most prominent entity of the context. 
The model for detecting this bias only uses contexts without questions.
\item \emph{Shallow}: we design a very shallow model to capture simple patterns of the dataset that may not be captured by the aforementioned biases. 
We use a simplified Bi-Directional Attention Flow (BiDAF) model \citep{seo2016bidirectional}
that uses 50-dimension Glove word embeddings, no character embeddings and a single layer of LSTM (instead of two). %
 
\squishend
For each examined dataset, we first automatically generate a biased dataset which only contains biased examples (eg: only examples with empty questions) for each individual bias type and split the resulting dataset into two halves. We then train a separate bias model for each half and use them to compute the bias weights of the other half.

\begin{table}[!htb]
\footnotesize
\centering
\resizebox{\columnwidth}{!}{%
\begin{tabular}{l|S[table-format=2.1]S[table-format=2.1]S[table-format=2.1]S[table-format=2.1]|S[table-format=2.1]S[table-format=2.1]}
\toprule
 \textbf{Dataset}  & \multicolumn{1}{c}{\textbf{wh.}} & \multicolumn{1}{c}{\textbf{emp.}} & \multicolumn{1}{c}{\textbf{lex.}} & \multicolumn{1}{c}{\textbf{shal.}} & \multicolumn{1}{|c}{\textbf{one}} & \multicolumn{1}{c}{\textbf{all}} \\ 
\midrule
SQuAD   &  17.9 & 8.8 & 51.9 & 32.7 & 61.9 & 3.4  \\
Hotpot  &  26.8 & 18.2 &  56.5 &  45.1 &  74.5 & 6.9 \\
Trivia  & 29.6 & 26.8 & 41.6 & 21.3 & 58.1 & 6.2   \\
News    &  16.2 & 7.9 & 11.4 & 17.4 & 31.8 & \bfseries 1.0  \\
NQ     &   47.5 &  38.5 & 51.0 & 38.7 & 64.8 &  \bfseries 23.2\\ 
\bottomrule
\end{tabular}
}
\caption{The ratio of examples that are answered correctly by the bias models. `\emph{one}' shows the ratio of examples that contain at least one bias. `\emph{all}' shows the ratio for examples that contain all biases.}
\label{table:bias_models}
\end{table}

\subsection{Data}
\label{sect:data}
We use five training datasets. This includes SQuAD \citep{Rajpurkar_2016}, HotpotQA \citep{yang-etal-2018-hotpotqa}, TriviaQA \citep{m2017triviaqa}, NewsQA \citep{trischler2016newsqa}, and Natural Questions (NQ) \citep{kwiatkowski-etal-2019-natural}.
For evaluating the out-of-domain generalization of models, we use six datasets. This includes BioASQ \citep{wiese-etal-2017-neural-question}, DROP \citep{dua-etal-2019-drop}, DuoRC \citep{saha-etal-2018-duorc}, RACE \citep{lai-etal-2017-race}, RelationExtraction \citep{levy-etal-2017-zero}, and TextbookQA \citep{8100054}. 
For all training and evaluation datasets, we use the version that are provided by the MRQA shared task, in which %
all examples can be solved using extractive answer selection. %
Detailed statistics of all datasets are reported in the appendix.

\begin{table*}[!htb]
\footnotesize
\centering
\begin{tabular}{lS[table-format=2.1]S[table-format=2.1]S[table-format=2.1]|S[table-format=2.1]S[table-format=2.1]S[table-format=2.1]}
\toprule
 &\multicolumn{3}{c|}{\textbf{NQ}} 
   &\multicolumn{3}{c}{\textbf{TriviaQA}} \\
 \textbf{Dataset} &\multicolumn{1}{c}{\textbf{Baseline}} & \multicolumn{1}{c}{\textbf{Mb-WL}} & \multicolumn{1}{c|}{\textbf{Mb-CR}}
 &\multicolumn{1}{c}{\textbf{Baseline}} & \multicolumn{1}{c}{\textbf{Mb-WL}} & \multicolumn{1}{c}{\textbf{Mb-CR}}\\ 
\midrule
Dev. & 63.66  & \bfseries 64.90  & 64.95                &58.24  &\bfseries 59.87  &59.09  \\ \midrule
\textbf{I-$\boldsymbol\Delta$} &   & +1.24  & +1.29     &       &1.63   &0.85   \\ \midrule
DROP   & 19.10 & \bfseries 21.76  & 21.29               &9.12   &\bfseries 9.51   &9.12   \\
RACE  & 20.47 & 22.85  & \bfseries 23.00                &15.58  &15.58  &\bfseries 15.88  \\
BioSQ  & 34.91  & 36.10  & \bfseries 36.44              &26.60  &28.13  &\bfseries 28.39   \\
TxtQA  & 30.94  & 33.87 & \bfseries 34.66               &17.76  &17.63  &\bfseries 17.9   \\
RelExt & 63.74 & 63.06  & \bfseries 64.01               &62.01  &61.46  &\bfseries 62.45  \\
DuoRC  & 34.78  & 36.64  & \bfseries 38.71              &24.32  &\bfseries 27.58  &26.58  \\ \midrule
AVG & 33.99  & 35.71  & \bfseries 36.35                 &25.90  &26.65  &\bfseries 26.72  \\ \midrule
\textbf{O-$\boldsymbol\Delta$} &   & +1.72  & +2.36      &       &0.75   &0.82   \\
\bottomrule
\end{tabular}
\caption{The impact of our debiasing framework in a single-domain training setting when the model is trained on NQ and TriviaQA. I-$\Delta$ and O-$\Delta$ are the average EM improvements on in-domain and out-of-domain experiments, respectively. %
Highest scores on each evaluation set are boldfaced.}
\label{tabel:sinlge_task}
\end{table*}

\subsection{Evaluation Settings}
We evaluate our proposed methods in two different settings: (1) \emph{single-domain} (SD)%
, and (2) \emph{multi-domain} (MD).
In SD, the model is trained on a single dataset.
For the MD setting, we use all the training datasets of \S\ref{sect:data}.
Our baseline within this setting is the multi-task model of \citet{fisch2019mrqa} which is a BERT model trained on all datasets with multi-task learning.
We refer to this baseline as \emph{MT-BERT}.

We report %
Exact Match (EM), i.e., whether the predicted answer exactly matches the correct one. 
We include the corresponding $F_1$ scores which measure the overlap rate between the predicted answer and the gold one in the appendix.

\section{Results}

\subsection{Strength of biases on different datasets}
\label{sec:bias_strength}

We report the ratio of the examples for each dataset that are correctly answered by our bias models (see \S\ref{sect:bias_models}) in Table~\ref{table:bias_models}. 
A higher ratio corresponds to a stronger observed bias. %
We observe that%
~(1) different datasets are more affected by certain biases,
e.g., the ratio of examples that can be answered without the question (the \emph{empty question} bias) is 8\% in SQuAD while it is 38\% in NQ, 
(2) NewsQA is least affected by biases overall while NQ and HotpotQA are most affected,
(3) only few instances are affected by all four biases,
and (4) except for NewsQA, the majority of training examples are affected by at least one bias. Therefore, methods that down-weight or ignore all biased examples will considerably weaken %
the overall training signal. %

\subsection{Impact of debiasing on SD training}
\label{sect:results-sd}

Table \ref{tabel:sinlge_task} shows the results of models trained on a \emph{single domain}.
We report the results when we train the model on NQ and TriviaQA, which have the highest and a medium percentage of examples that contain all biases (according to the \textbf{all} column in Table~\ref{table:bias_models}), 
respectively.
The results of SD based on other training datasets are reported in the appendix.

We observe that (1) without using any additional training examples or increasing the model size, we can improve generalization by using our debiasing methods, (2) the impact of debiasing methods is stronger when the training data is more biased, and (3) the use of our proposed debiasing methods not only improve generalization, but it also improves the performance on the in-domain evaluation dataset, which contains similar biases as those of the training data. This is in contrast to previous work that either decreases the in-domain performance \citep{he-etal-2019-unlearn,Clark_2019,mahabadiend}, or at most preserves it
\citep{utama2020mind}. We analyze the reason for this in \S\ref{sect:why_improves}.

\subsection{Impact of debiasing on MD training}
\label{sect:md_results}

Table \ref{tabel:multi_task} shows the results of the \emph{multi-domain} setting.
\citet{Talmor_2019} show that training \emph{MT-BERT} on multiple domains leads to robust generalization.
Since \emph{MT-BERT} is trained on multiple domains simultaneously, which are not equally affected by different biases, %
the model is less likely to learn these patterns.
However, our results show that our debiasing methods further improve the average EM scores 
by more than one point even if the model is trained on multiple domains.

\begin{table}[!htb]
\footnotesize
\centering
\begin{tabular}{lS[table-format=2.1]S[table-format=2.1]S[table-format=2.1]}
\toprule
  \textbf{Dataset}&\multicolumn{1}{c}{\textbf{MT-BERT}} & \multicolumn{1}{c}{\textbf{Mb-WL}} & \multicolumn{1}{c}{\textbf{Mb-CR}} \\ 
\midrule
SQuAD  &77.52  & \bfseries 79.87 & 79.59    \\
Hotpot           &58.77 & 59.43  & \bfseries 59.58   \\
Trivia        & \bfseries 63.66    & 62.5    & 62.94     \\
News              & 45.96   & 49.36 & \bfseries 49.72  \\
NQ                  & 64.86     & \bfseries 65.52  & 65.5     \\ \midrule
I-AVG               & 62.15      & 63.34  & \bfseries 63.47   \\  \midrule
\textbf{I-$\boldsymbol\Delta$}&      & +1.18 & +1.31   \\  \midrule
DROP                & \bfseries29.34   & 29.27  & 28.14     \\
RACE                & \bfseries 30.86   & 30.12  & 29.82    \\
BioSQ              & 46.94    & 49.6   & \bfseries50.2    \\
TxtQA          & 39.06    & 43.38  & \bfseries 44.58     \\
RelExt              & \bfseries 73.93   & 73.64  & 72.96  \\
DuoRC               & 44.37   & \bfseries 46.17  & 45.64   \\  \midrule
O-AVG        & 44.08     & \bfseries 45.36  & 45.22   \\   \midrule
\textbf{O-$\boldsymbol\Delta$}&        & +1.28  & +1.14     \\
\bottomrule
\end{tabular}
\caption{Impact of our debiasing methods when trained on multiple domains. \emph{MT-BERT} is trained with the MRQA setup. The upper and bottom block present the in-domain and out-of-domain scores, respectively.} %
\label{tabel:multi_task}
\end{table}

\section{Discussion and Analysis}
In this section, we discuss the benefits and limitations of our framework. 
\subsection{Why our debiasing improves in-domain and out-of-domain performances?}
\label{sect:why_improves}
The main differences of our proposed framework to the state-of-the-art debiasing approaches are as follows:
\begin{itemize} 
	\item It is a general framework and can be used with any bias-aware training objectives, e.g., that of \citet{utama2020mind} or \citet{mahabadiend}.
	\item It models multiple biases at the same time compared to \citet{utama2020mind}'s confidence-regularization method.
	\item It incorporates both dataset-level and example-level weights for each bias, and combines them using the multi-bias weighting function, while \citet{mahabadiend}'s DFL method simply average example-level weights of different biases. %
\end{itemize}

\citet{utama2020mind}'s %
CR method can be modeled in our \emph{Mb-CR} method by only modeling a single bias and removing the $F_B(x_i)$ combination function.

\citet{mahabadiend} propose two different methods among which the Debiased Focal Loss (DFL) approach has a better performance. Therefore, we use \emph{DFL} in our comparisons.

The comparison of our methods vs.\ (1)  \citet{utama2020mind}'s \emph{CR} will indicate whether modeling multiple biases at once is a key factor on the resulting improvements, and (2) \citet{mahabadiend}'s \emph{DFL} will indicate whether our proposed 
methods for modeling of multiple biases  
improves the performance or any method that models multiple biases jointly will have the same impact.
For a fair comparison, we use the same bias types and bias weights in all the debiasing methods.

\begin{table}[!htb]
\footnotesize
\centering
\resizebox{\columnwidth}{!}{
\begin{tabular}{llllll}
\toprule
 \multicolumn{2}{c}{} &\multicolumn{2}{c}{\textbf{in-domain}} & \multicolumn{2}{c}{\textbf{out-of-domain}} \\
 \multicolumn{2}{c}{\textbf{Method}}
 &\textbf{EM}  &\textbf{I-$\boldsymbol\Delta$} &\textbf{EM}     &\textbf{O-$\boldsymbol\Delta$}  \\
\midrule
\multirow{4}{*}{SD} 
& Baseline  &63.66     & -            &33.99   & -              \\
& CR(lex.)    & 58.32 & -5.34 & 34.28 & 0.29           \\
& DFL       &64.32     & +0.66             &34.35   & +0.36 \\ 
& \bfseries Mb-CR     &64.95     & \bfseries +1.29    &36.35  & \bfseries +2.36  \\ 

\midrule
\multirow{4}{*}{MD} 
& Baseline  &62.15     & -                 &44.08   & -                \\
& CR(lex.)    &  61.35 & -0.80 & 43.70 & -0.39           \\
& DFL       &63.35     & +1.20             &44.44   & +0.36            \\
& \bfseries Mb-CR     &63.47     & \bfseries +1.31   &45.22   & \bfseries +1.14  \\

\bottomrule
\end{tabular}
}
\caption{Comparisons with \citet{utama2020mind} and \citet{mahabadiend} debiasing methods, i.e., \emph{CR(lex.)} and \emph{DFL}, respectively.}
\label{tabel:comparisons}
\end{table}

Table~\ref{tabel:comparisons} presents the corresponding EM 
scores of these experiments. %
For SD experiments, we use NQ for training since it contains the largest number of training examples.
For the CR method of \citet{utama2020mind} that handles a single bias, we use the \emph{lexical overlap} bias, as it is the most dominant bias in the majority of our training datasets (see Table~\ref{table:bias_models}).\footnote{The results of CR with other bias types, i.e., \emph{Mb-CR} with a single bias, is reported in Table~\ref{tabel:ablation_results_2}.}

Based on the SD results, we observe that (1) debiasing only based on the \emph{lexical overlap} bias, which is the strongest bias in the training data, considerably drops the in-domain performance, and it has a negligible impact on out-of-domain results, and (2) while combining all biases using \emph{DFL} improves the in-domain results, it does not have a significant impact on out-of-domain performances.
This shows %
the importance of (a) concurrent modeling of multiple-biases, and (b) our proposed multi-bias methods in improving the overall performance. %
We will further investigate the impact of each of the components in our framework in \S\ref{sect:ablation}.

The results of \emph{CR(lex)} in the MD setting show that debiasing based on a single bias---one that  is common in most of training datasets---negatively impacts the in-domain and out-of-domain performances.
Similar to the SD results, the \emph{DFL} bias combination 
has a more positive impact on in-domain instead of out-of-domain in MD results.

Overall, both SD and MD results show the effectiveness of our proposed framework for both in-domain and out-of-domain setups.

\subsection{Impact of the Framework Components}
\label{sect:ablation}
We investigate the impact of the components of our framework including: (1) knowledge distillation (KD): by replacing the teacher probabilities with gold labels in \emph{Mb-WL}; and (2) the scaling factor ($F_S$): by removing the scaling factor from Equation~\ref{eq:combine}.
Table \ref{tabel:ablation_results_1} reports the results for the SD setting when the model is trained on NQ.
The results show that KD has a big impact on the generalization of \emph{Mb-WL}, while $F_S$ has a stronger impact on \emph{Mb-CR}'s generalization.

\begin{table}[!htb]
\footnotesize
\centering
\begin{tabular}{lll|ll}
\toprule
 &\multicolumn{2}{c}{\textbf{Mb-WL}} & \multicolumn{2}{c}{\textbf{Mb-CR}} \\
&\multicolumn{1}{l}{\textbf{I-$\boldsymbol\Delta$}} & \multicolumn{1}{l}{\textbf{O-$\boldsymbol\Delta$}} &\multicolumn{1}{l}{\textbf{I-$\boldsymbol\Delta$}} & \multicolumn{1}{l}{\textbf{O-$\boldsymbol\Delta$}}   \\ 
\midrule
no KD  &-0.60    &-1.95 &- &- \\
no $F_S$  &-0.32 &+0.38 &-0.57 &-1.34\\
\bottomrule
\end{tabular}
\caption{Impact of knowledge distillation and the scaling factor in our \emph{Mb-WL} and \emph{Mb-CR} methods. }
\label{tabel:ablation_results_1}
\end{table}
In addition, we evaluate the impact of combining multiple biases in Table \ref{tabel:ablation_results_2} by using a single bias at a time instead of modeling multiple biases.
The results show that multi-bias modeling (1) is more useful than modeling any individual bias %
for both in-domain and out-of-domain experiments, and (2) has a more significant impact on \emph{Mb-CR} compared to \emph{Mb-WL}.   

\begin{table}[!htb]
\footnotesize
\centering
\begin{tabular}{lll|ll}
\toprule
 &\multicolumn{2}{c}{\textbf{Mb-WL}} & \multicolumn{2}{c}{\textbf{Mb-CR}} \\
&\multicolumn{1}{l}{\textbf{I-$\boldsymbol\Delta$}} & \multicolumn{1}{l}{\textbf{O-$\boldsymbol\Delta$}} &\multicolumn{1}{l}{\textbf{I-$\boldsymbol\Delta$}} & \multicolumn{1}{l}{\textbf{O-$\boldsymbol\Delta$}}   \\ 
\midrule
wh. only  &-0.75      &-1.01 &-3.88 &-0.8\\
emp. only   &-0.13      &-0.95 &-2.24 &-0.39\\
lex. only  &-0.66      &-0.69 &-6.63 &-2.07\\
shal. only  &+0.46      &-0.68 &-4.11 &-0.86\\
\bottomrule
\end{tabular}
\caption{The performance differences between using single-bias modeling compared to multi-bias modeling. All models are trained on NQ dataset.}
\label{tabel:ablation_results_2}
\end{table}

\subsection{Is debiasing always beneficial?}
We hypothesize that applying debiasing methods will not lead to performance gains if (1) the presence of examined biases is not strong in the training data, i.e., if most of the examples are unbiased, and therefore the model that is trained on this data will not be biased, to begin with,
and (2) %
the out-of-domain
set strongly contain the biases 
based on which the model is debiased during training.%

To verify the first hypothesis, we evaluate the single-domain experiments using the NewsQA dataset that contains the smallest ratio of biased examples, i.e., only 1\% of the data contain all of the examined biases.
The results are reported in Table~\ref{tabel:sinlge_task_newsq_short}, which in turn confirms our hypothesis.

\begin{table}[!htb]
\footnotesize
\centering
\begin{tabular}{lS[table-format=2.1]S[table-format=2.1]}
\toprule
\textbf{Dataset} & \multicolumn{1}{c}{\textbf{Mb-WL}}& \multicolumn{1}{c}{\textbf{Mb-CR}} \\ 
\midrule
\textbf{I-$\boldsymbol\Delta$}  & -0.26  & 0.14  \\ \midrule
\textbf{O-$\boldsymbol\Delta$} & 0.49  & -0.10 \\
\bottomrule
\end{tabular}
\caption{Impact of our methods when trained on NewsQA that contains few biased examples. }
\label{tabel:sinlge_task_newsq_short}
\end{table}

Regarding the second hypothesis, we report the results of the bias models on the evaluation sets in Table~\ref{tabel:ood_bias_results}.
The results of all bias models are very high on RelExt compared to other evaluation datasets, and as we see from the results of both SD and MD settings in Table~\ref{tabel:sinlge_task} and ~\ref{tabel:multi_task}, our debiasing methods are the least effective on improving the out-of-domain performance on this evaluation set.

\begin{table}[!htb]
\footnotesize
\centering
\begin{tabular}{l|S[table-format=2.1]S[table-format=2.1]S[table-format=2.1]S[table-format=2.1]}
\toprule
 \textbf{Dataset}  & \multicolumn{1}{r}{\textbf{wh.}} & \multicolumn{1}{r}{\textbf{emp.}} & \multicolumn{1}{r}{\textbf{lex.}} & \multicolumn{1}{r}{\textbf{shal.}}  \\ 
\midrule
DROP   &8.98   &5.06   &14.64  &2.99 \\
RACE   &7.42   &3.56   &15.13  &2.67 \\
BioSQ  &12.70   &10.44  &25.86  &5.12 \\
TxtQA  &8.65   &5.46   &15.44  &3.93 \\
RelExt & \bfseries 30.16  &\bfseries 21.13  &\bfseries 57.56  &\bfseries 19.67 \\
DuoRC  &5.67   &2.93   &24.52  &4.73 \\
\bottomrule
\end{tabular}
\caption{The EM scores of the bias models, which are trained on NQ, on out-of-domain evaluation sets.}
\label{tabel:ood_bias_results}
\end{table}

\section{Conclusion}
In this paper we (1) investigate the impact of debiasing methods on QA model generalization for both single and multi-domain training scenarios, and (2) propose a new framework for improving the in-domain and out-of-domain performances by concurrent modeling of multiple biases. Our framework weights each training example according to multiple biases and based on the strength of each bias in the training data.
It uses the resulting bias weights in the training objective to prevent the model from mainly focusing on learning biases.
We evaluate our framework using two different training objectives, i.e., multi-bias confidence regularization and multi-bias loss re-weighting, and show its effectiveness in both single and multi-domain training scenarios.
We further compare our framework with
two state-of-the-art debiasing methods of \citet{utama2020mind} and \citet{mahabadiend}.
We show that knowledge distillation, modeling multiple biases at once, and weighting the impact of each bias based on its strength in the training data are all important factors in improving the in-domain and out-of-domain performances.
While recent literature on debiasing in NLP focuses on improving the performance on %
adversarial evaluation sets,
this work opens new research directions on wider uses of debiasing methods.
The main advantage of using our debiasing methods is that they improve the performance and generalization without requiring additional training data or larger models. 
Future work could build upon our framework by applying it to a wide range of tasks beyond QA using task-specific bias models.

\section*{Acknowledgements}

This work was supported by
the German Federal Ministry of Education and Research (BMBF), the German Research Foundation under grant EC 503/1-1 and GU 798/21-1, and
the Hessen State Ministry for Higher Education, Research and the Arts within their joint support of
the National Research Center for Applied Cybersecurity ATHENE.
The first author of the paper is supported by a scholarship from Technical University of Darmstadt.
The authors would like to thank Prasetya Ajie Utama, Jan-Christoph Klie, Benjamin Schiller, Gisela Vallejo, Rabeeh Karimi Mahabadi, and the anonymous reviewers for their valuable feedbacks.

\bibliography{main}
\bibliographystyle{acl_natbib}

\newpage
\appendix 
\section{Training details}
We use the same hyperparameters as the MRQA shared task. To be more specific, we use BertAdam optimizer with a learning rate of $3 \times 10^{-5}$ and batch size of 6. We sample all training examples in each dataset during training and evaluation. All our models are trained for 2 epochs. We choose the size of 512 tokens to be the maximum sequence fed into the neural network. Contexts with longer tokens will be split into several training instances. The single domain experiment takes roughly 3 hours on a single Nvidia Tesla V100-SXM3-32GB GPU while it takes around 15 hours for the multi-domain experiment on the same GPU.

\section{Dataset statistics}
Table~\ref{tabel:dataset_statistics} presents a brief description for each of the examined training and evaluation sets.
\begin{table*}[!htb]
\footnotesize
\centering
\begin{tabular}{lllrrrr}
\toprule
 \textbf{Dataset}  & \multicolumn{1}{c}{\textbf{Question (Q)}} & \multicolumn{1}{c}{\textbf{Context (C)}} & \multicolumn{1}{c}{$\|\textbf{Q}\|$} & \multicolumn{1}{c}{$\|\textbf{C}\|$} & \multicolumn{1}{c}{\textbf{train}} & \multicolumn{1}{c}{\textbf{dev}} \\ 
\midrule
SQuAD   &Crowdsourced &Wikipedia &11 &137 &86,588 &10,507  \\
Hotpot  &Crowdsourced &Wikipedia &22 &232 &72,928 &5,904  \\
Trivia  &Trivia    &Web snippets &16 &784 &61,688 &7,785  \\
News    &Crowdsourced &News articles &8  &599 &74,160 &4,212 \\
NQ      &Search logs  &Wikipedia &9  &153  &104,071 &12,836 \\ 
\midrule
DROP  &Crowdsourced &Wikipedia  &11 &243  &- &1,503  \\
RACE  &Domain experts &Examinations &12 &349 &- &674 \\
BioSQ &Domain experts &Science article &11 &248 &- &1,504 \\
TxtQA &Domain experts &Textbook &11 &657 &- &1,503  \\
RelExt &Synthetic &Wikipedia &9 &30 &-  &2,948 \\ 
DuoRC  &Crowdsourced &Movie plots &9 &681 &- &1,501 \\
\bottomrule
\end{tabular}
\caption{The detailed statistics about the datasets. The upper block shows five domains used for training, the lower block shows six domains used for evaluation. $\|\textbf{Q}\|$ and $\|\textbf{C}\|$ denotes the average token length in Question and Context, respectively. The \textbf{train} and \textbf{dev} columns show the numbers of examples in the corresponding training and development sets, respectively.}
\label{tabel:dataset_statistics}
\end{table*}

\section{SD results using other training data}
We report the results of the SD setting using NQ, TriviaQA, and NewsQA in the paper.
Table~\ref{table:sd_others_em} reports the results, using the EM score, on the remaining training data, i.e., SQuAD and HotpotQA.
Debiasing the model on SQuAD has a more positive impact on out-of-domain results while debiasing the model that is trained on HotpotQA has a better impact on in-domain performances.

\section{Results using F1 scores}
The results in the paper are reported using the EM score. Table~\ref{tabel:sinlge_task_f1}-Table~\ref{tabel:ood_bias_results_f1} show the results of this work using F1 scores.
The main difference of EM and F1 scores are for answers whose corresponding span contains more than one word. If a system partially detects the correct span boundary, it receive a partial F1 score but a zero EM score. 
As we see, the findings of the paper would remain the same using F1 scores instead of EM scores.

\begin{table*}[!htb]
\footnotesize
\centering
\begin{tabular}{lS[table-format=2.1]S[table-format=2.1]S[table-format=2.1]|S[table-format=2.1]S[table-format=2.1]S[table-format=2.1]}
\toprule
 &\multicolumn{3}{c|}{\textbf{SQuAD}} 
  &\multicolumn{3}{c}{\textbf{HotpotQA}}  \\
 &\multicolumn{1}{c}{\textbf{Baseline}} & \multicolumn{1}{c}{\textbf{Mb-WL}} & \multicolumn{1}{c|}{\textbf{Mb-CR}}
 &\multicolumn{1}{c}{\textbf{Baseline}} & \multicolumn{1}{c}{\textbf{Mb-WL}} & \multicolumn{1}{c}{\textbf{Mb-CR}} \\ 
 \textbf{Dataset} & \textbf{EM}   &\textbf{EM}    &\textbf{EM} & \textbf{EM}   &\textbf{EM}    &\textbf{EM} \\
\midrule
dev.                           &79.24   &79.82  &79.39  &55.48  &56.48  &56.47   \\ 
\midrule
\textbf{I-$\boldsymbol\Delta$} &        &0.58   &0.15   &       &1.00   &0.99   \\
\midrule
DROP                           &17.30   &16.9   &18.9   &19.69  &20.83  &19.43  \\
RACE                           &23.59   &24.18  &25.07  &17.51  &16.77  &17.95   \\
BioSQ                          &45.28   &44.02  &42.49  &37.90  &37.96  &37.5    \\
TxtQA                          &33.67   &36.19  &36.06  &14.97  &15.97  &16.1   \\
RelExt                         &68.93   &68.42  &68.15  &63.06  &60.89  &61.67   \\
DuoRC                          &40.57   &43.77  &43.24  &28.78  &32.91  &31.65   \\
\midrule
AVG                            &38.22   &38.91  &38.99  &30.32  &30.89  &30.72   \\
\midrule
\textbf{O-$\boldsymbol\Delta$} &        &0.69   &0.76         &  &0.57   &0.40    \\
\bottomrule
\end{tabular}
\caption{The impact of our debiasing methods on SQuAD and HotpotQA. I-$\Delta$ and O-$\Delta$ indicate the average improvements in in-domain and out-of-domain experiments, respectively.}
\label{table:sd_others_em}
\end{table*}

\begin{table*}[!htb]
\footnotesize
\centering
\begin{tabular}{lS[table-format=2.1]S[table-format=2.1]S[table-format=2.1]|S[table-format=2.1]S[table-format=2.1]S[table-format=2.1]}
\toprule
 &\multicolumn{3}{c|}{\textbf{NQ}} 
   &\multicolumn{3}{c}{\textbf{TriviaQA}} \\
\textbf{Dataset} &\multicolumn{1}{c}{\textbf{Baseline}} & \multicolumn{1}{c}{\textbf{Mb-WL}} & \multicolumn{1}{c|}{\textbf{Mb-CR}}
 &\multicolumn{1}{c}{\textbf{Baseline}} & \multicolumn{1}{c}{\textbf{Mb-WL}} & \multicolumn{1}{c}{\textbf{Mb-CR}}\\ 
\midrule
Dev.                            &75.36  &76.44  &76.57  &64.66  &66.44  &66.08    \\ \midrule
\textbf{I-$\boldsymbol\Delta$}  & &1.08   &1.21   &       &1.78   &1.42   \\ \midrule
DROP                            &28.75  &31.41  &30.93  &14.89  &15.2   &14.02    \\
RACE                            &30.04  &32.1   &33.03  &22.15  &21.77  &22.06    \\
BioSQ                           &52.13  &54.46  &53.18  &36.68  &39.94  &40.98    \\
TxtQA                           &40.03  &43.03  &43.48  &21.86  &21.75  &21.94    \\
RelExt                          &77.68  &77.45  &77.75  &73.86  &73.09  &74.3     \\
DuoRC                           &43.44  &45.37  &47.04  &31.48  &34.64  &33.7     \\  \midrule
AVG                             &45.35  &47.30  &47.57  &33.49  &34.40  &34.50    \\  \midrule
\textbf{O-$\boldsymbol\Delta$}  &       &1.96   &2.22   &       &0.91   &1.01     \\
\bottomrule
\end{tabular}
\caption{The impact of our debiasing framework in a single-domain training setting when the model is trained on NQ and TriviaQA. I-$\Delta$ and O-$\Delta$ are the average improvements on in-domain and out-of-domain experiments, respectively. Results are reported based on F1 scores. } %
\label{tabel:sinlge_task_f1}
\end{table*}
\begin{table*}[!htb]
\footnotesize
\centering
\begin{tabular}{lS[table-format=2.1]S[table-format=2.1]S[table-format=2.1]|S[table-format=2.1]S[table-format=2.1]S[table-format=2.1]}
\toprule
 &\multicolumn{3}{c|}{\textbf{SQuAD}} 
  &\multicolumn{3}{c}{\textbf{HotpotQA}}  \\
 &\multicolumn{1}{c}{\textbf{Baseline}} & \multicolumn{1}{c}{\textbf{Mb-WL}} & \multicolumn{1}{c|}{\textbf{Mb-CR}}
 &\multicolumn{1}{c}{\textbf{Baseline}} & \multicolumn{1}{c}{\textbf{Mb-WL}} & \multicolumn{1}{c}{\textbf{Mb-CR}} \\ 
 \textbf{Dataset} & \textbf{F1}   &\textbf{F1}    &\textbf{F1} & \textbf{F1}   &\textbf{F1}    &\textbf{F1} \\
\midrule
dev.                           & 86.93  &86.99  &86.72  &73.24  &73.52  &73.47   \\ \midrule
\textbf{I-$\boldsymbol\Delta$} &        &0.06   &-0.21  &       &0.28   &0.23       \\ \midrule
DROP                           & 24.52  &24.23  &26.3   &30.62  &31.68  &30.73       \\
RACE                           & 34.95  &35.67  &36.21  &26.44  &26.6   &26.65        \\
BioSQ                          & 57.36  &56.33  &54.8   &52.31  &52.33  &52.72        \\
TxtQA                          & 41.48  &44.01  &43.68  &22.52  &21.68  &22.59       \\
RelExt                         & 80.51  &80.19  &80.33  &76.60  &73.75  &74.84        \\
DuoRC                          & 49.10  &51.35  &50.97  &37.67  &41.63  &40.22        \\ \midrule
AVG                            & 47.99  &48.63  &48.72  &41.03  &41.28  &41.29        \\ \midrule
 \textbf{O-$\boldsymbol\Delta$} &        &0.64   &0.73   &       &0.25   &0.26         \\
\bottomrule
\end{tabular}
\caption{The impact of our debiasing methods on SQuAD and HotpotQA based on F1 scores. I-$\Delta$ and O-$\Delta$ indicate the average improvements in in-domain and out-of-domain experiments, respectively.}
\label{table:sd_others_f1}
\end{table*}

\begin{table}[!htb]
\footnotesize
\centering
\begin{tabular}{lS[table-format=2.1]S[table-format=2.1]S[table-format=2.1]}
\toprule
   &\multicolumn{1}{c}{\textbf{MT-BERT}} & \multicolumn{1}{c}{\textbf{Mb-WL}} & \multicolumn{1}{c}{\textbf{Mb-CR}} \\ 
\midrule
SQuAD   &85.78  &87.25  &87.26               \\
Hotpot  &75.52  &76.47  &76.46            \\
Trivia  &69.48  &69.6   &69.89            \\
News    &61.39  &64.49  &64.84            \\
NQ      &76.8   &77.28  &77.23                  \\ \midrule
I-AVG   &73.79  &75.02  &75.14                  \\  \midrule
\textbf{I-$\boldsymbol\Delta$}        & &1.22   &1.34 \\ \midrule
DROP    &37.83  &37.71  &36.61   \\
RACE    &41.21  &40.96  &40.64   \\
BioSQ   &62.28  &64.16  &64.53   \\
TxtQA   &47.40  &51.71  &52.93   \\
RelExt  &84.10  &84.45  &84.03   \\
DuoRC   &53.33  &55.16  &54.33   \\  \midrule
O-AVG          &54.36   &55.69  &55.51 \\ \midrule
\textbf{O-$\boldsymbol\Delta$}        & &1.33   &1.15 \\ 
\bottomrule
\end{tabular}
\caption{F1 scores of our debiasing methods when trained on multiple domains. \emph{MT-BERT} is the MRQA baseline trained on five training datasets.} 
\label{tabel:multi_task_f1}
\end{table}

\begin{table}[!htb]
\footnotesize
\centering
\resizebox{\columnwidth}{!}{
\begin{tabular}{llllll}
\toprule
 \multicolumn{2}{c}{} &\multicolumn{2}{c}{\textbf{in-domain}} & \multicolumn{2}{c}{\textbf{out-of-domain}} \\
 \multicolumn{2}{c}{\textbf{Method}}
 &\textbf{F1}  &\textbf{I-$\boldsymbol\Delta$} &\textbf{F1}     &\textbf{O-$\boldsymbol\Delta$}  \\
\midrule
\multirow{4}{*}{SD} 
& Baseline  & 75.36       & -              &45.35 &-             \\
& CR(lex.)    & 71.32 & -4.04 & 46.05 & 0.70           \\
& DFL       &75.97     & +0.61             &45.90   & +0.55 \\ 
& \bfseries Mb-CR     &76.57     & \bfseries +1.21    &47.57  & \bfseries +2.22  \\ 
\midrule
\multirow{4}{*}{MD} 
& Baseline  &73.79     &-                  &54.36   &-               \\
& CR(lex.)    &  73.15 & -0.65 & 54.45 & -0.09           \\
& DFL       &74.93     & +1.13             &55.54   & +1.18            \\
& \bfseries Mb-CR     &75.18     & \bfseries +1.38   &55.68   & \bfseries +1.32  \\
\bottomrule
\end{tabular}
}
\caption{Comparisons with \citet{utama2020mind} and \citet{mahabadiend} debiasing methods, i.e., \emph{CR(lex.)} and \emph{DFL}, respectively. F1 scores reported.}
\label{tabel:comparisons_f1}
\end{table}

\begin{table}[!htb]
\footnotesize
\centering
\begin{tabular}{lll|ll}
\toprule
 &\multicolumn{2}{c}{\textbf{Mb-WL}} & \multicolumn{2}{c}{\textbf{Mb-CR}} \\
&\multicolumn{1}{l}{\textbf{I-$\boldsymbol\Delta$}} & \multicolumn{1}{l}{\textbf{O-$\boldsymbol\Delta$}} &\multicolumn{1}{l}{\textbf{I-$\boldsymbol\Delta$}} & \multicolumn{1}{l}{\textbf{O-$\boldsymbol\Delta$}}   \\ 
\midrule
                & 76.44 &47.30  &76.57  &47.57  \\      \midrule
no KD           & -0.43 &-1.34  &-       &-       \\
no $F_S$           & -0.30 &-0.16  &-0.48  &-1.00  \\
wh. only          & -0.52 &-0.97  &-3.13  &-0.63  \\
emp. only        & -0.20 &-1.12  &-1.83  &-0.30  \\
lex. only        & -0.55 &-0.53  &-5.25  &-1.52  \\
shal. only       & 0.25  &-0.61  &-2.86  &-1.15  \\
\bottomrule
\end{tabular}
\caption{F1 scores for different variations of the \emph{Mb-WL} debiasing method. $F_S$ refers to scaling factor. 
}
\label{tabel:ablation_results_f1}
\end{table}

\begin{table}[!htb]
\footnotesize
\centering
\begin{tabular}{lS[table-format=2.1]S[table-format=2.1]S[table-format=2.1]}
\toprule
\textbf{Dataset} &\multicolumn{1}{c}{\textbf{Baseline}} & \multicolumn{1}{c}{\textbf{Mb-WL}}& \multicolumn{1}{c}{\textbf{Mb-CR}} \\ 
\midrule
Dev. & 50.31 & 50.05  & \bfseries 50.45 \\ \midrule
\textbf{I-$\boldsymbol\Delta$} &  & -0.26  & 0.14  \\ \midrule
DROP   & \bfseries 13.51 & 12.71 & 12.71  \\
RACE  & \bfseries 23.00  & 22.55  & 20.92  \\
BioSQ  & 31.52  & \bfseries 33.11 & \bfseries 33.11  \\
TxtQA  & 28.94  & \bfseries 31.07  & 30.54  \\
RelExt & \bfseries 50.88 & 50.75  & 50.58  \\
DuoRC  & 36.18  & \bfseries 36.78  & 35.58  \\ \midrule
AVG & 30.67  & 31.16  & 30.57  \\ \midrule
\textbf{O-$\boldsymbol\Delta$} &  & 0.49  & -0.10 \\
\bottomrule
\end{tabular}
\caption{The impact of debiasing methods evaluated using F1 scores when the model is trained on NewsQA that contains few biased examples. }
\label{tabel:sinlge_task_newsq}
\end{table}

\begin{table}[!htb]
\footnotesize
\centering
\begin{tabular}{l|S[table-format=2.1]S[table-format=2.1]S[table-format=2.1]S[table-format=2.1]}
\toprule
 \textbf{Dataset}  & \multicolumn{1}{r}{\textbf{wh.}} & \multicolumn{1}{r}{\textbf{emp.}} & \multicolumn{1}{r}{\textbf{lex.}} & \multicolumn{1}{r}{\textbf{shal.}}  \\ 
\midrule
DROP    &15.16  &8.19   &21.7   &8.61   \\
RACE    &13.16  &6.12   &23.09  &6.7    \\
BioSQ   &23.87  &18.73  &40.46  &13.12  \\
TxtQA   &12.69  &8.67   &21.18  &6.69   \\
RelExt  &\bfseries 41.78  &\bfseries 29.25  &\bfseries 71.88  &\bfseries 31.12  \\
DuoRC   &8.54   &4.23   &32.7   &9.26   \\
\bottomrule
\end{tabular}
\caption{The F1 scores of the bias models, which are trained on NQ, on evaluation sets.}
\label{tabel:ood_bias_results_f1}
\end{table}

\end{document}